\documentclass[conference,onecolumn,letterpaper,11pt]{IEEEtran}
\usepackage[letterpaper,margin=1in]{geometry}
\usepackage[utf8]{inputenc} 
\usepackage[T1]{fontenc}
\usepackage{url}
\usepackage{ifthen}
\usepackage{cite}
\usepackage{amssymb}
\usepackage[cmex10]{amsmath} %
\usepackage{algorithm}
\usepackage{algpseudocode}  %

\usepackage{algpseudocode} %

\usepackage{bm}       %
\usepackage{mathtools}%

\usepackage{bbm} %
\usepackage{multirow}
\usepackage{amsthm}

\theoremstyle{plain}
\newtheorem{theorem}{Theorem}

\theoremstyle{definition}
\newtheorem{assumption}{Assumption}

\theoremstyle{remark}

\usepackage{xcolor}
\usepackage{graphicx}
\usepackage{subcaption}
 \usepackage{booktabs}

\begin{document}
\title{Server-Side Sparse Assignment for Federated Mixture-of-Experts} 

\author{\IEEEauthorblockN{Xue Zheng and Aylin Yener}
\IEEEauthorblockA{Department of Electrical and Computer Engineering \\
The Ohio State University, Columbus, OH, USA\\
zheng.1822@osu.edu yener@ece.osu.edu}
}

\maketitle

\noindent\copyright\ 2026 IEEE. Personal use of this material is permitted.
Permission from IEEE must be obtained for all other uses, in any current or
future media, including reprinting/republishing this material for advertising
or promotional purposes, creating new collective works, for resale or
redistribution to servers or lists, or reuse of any copyrighted component
of this work in other works.\par\medskip

\begin{abstract}
    Large language models (LLMs) are increasingly recognized for their remarkable capabilities; federated learning (FL) offers a promising framework for collaboratively fine-tuning LLMs on edge devices while achieving data privacy. However, edge devices are inherently heterogeneous: clients may exhibit diverse data distributions and resource constraints, including limited memory and computation budgets. Recent federated Mixture-of-Experts (MoE) approaches address these challenges by using sparse expert activation or partial expert loading. However, existing partial-expert methods rely on heuristic full-pool observations or prioritize system-wide load balance, without guarantees for recovering client-specific relevant experts from partial observations. We propose \textbf{$S^3$-MoE}, a server-side sparse expert assignment framework that estimates client-specific expert relevance from partial observations and refines assignments through an exploit--explore mechanism. We also establish relevance-estimation error and exact Top-$K$ recovery guarantees. Experiments using a CNN-based MoE on CIFAR-100 show that $S^3$-MoE recovers task-dependent client--expert preferences and reduces conflicts caused by correlated expert selection across clients. Compared with Full-20 FedAvg, which gives each client access to all 20 experts for local Top-5 routing, $S^3$-MoE achieves 1.31 percentage points higher mean final accuracy while loading only $5$ experts per client.
\end{abstract}

\section{Introduction}
Federated learning (FL) ~\cite{mcmahan2017communication} is increasingly considered a promising framework for deploying large language models (LLMs) on edge devices while preserving data privacy. Recent advances in LLMs increasingly exploit sparsely activated
Mixture-of-Experts (MoE) architectures, as exemplified by Mixtral,
OLMoE, DeepSeek-V3, and Qwen3~\cite{jiang2024mixtral,
muennighoff2025olmoe,deepseekai2024deepseekv3,yang2025qwen3}. However, deploying such models on heterogeneous edge devices remains fundamentally challenging due to limited memory, computation, and communication resources. In practice, many edge devices cannot store or access the full model or expert pool during \textit{local fine-tuning}. 

Several approaches have been proposed to enable efficient on-device deployment. 
Parameter-efficient fine-tuning methods such as LoRA and local adapters reduce training overhead but still require loading the full backbone model locally~\cite{hu2021lora,fedalora2025,fedlora2024}. 
Split learning reduces local memory usage but introduces substantial communication overhead due to repeated transmission of intermediate activations~\cite{liu2024splitfl}. 
In contrast, Mixture-of-Experts (MoE) architectures offer a different form of efficiency by activating only a small subset of experts for each input~\cite{shazeer2017outrageously}. In practice, edge devices also face heterogeneous task-dependent workloads. For example, mobile phones may focus on conversational assistants, office computers on productivity tasks, and autonomous vehicles on speech and navigation functions. Due to limited local memory and computation budgets, many devices cannot store or fine-tune full MoE models locally. This motivates a promising direction for edge deployment: instead of loading the full model, fine-tune task-relevant expert subsets and reuse the learned preferences to deploy compact, personalized sub-MoEs for fully local inference.

However, existing federated MoE fine-tuning approaches primarily focus on sparse activation while still assuming full expert visibility~\cite{reisser2021fedmix,dun2023fedjets,yi2024pfedmoe,luo2024pfedmoap}, without considering memory constraints and task-specific device settings. Although some recent works explore partial expert loading, FedMoE assigns experts based on early-stage usage similarity, while FLEX-MoE uses client--expert performance feedback and an optimization-based allocation algorithm to balance expert utilization.
However, FedMoE relies on heuristic early-stage routing signals, while FLEX-MoE primarily focuses on load balancing; neither provides theoretical guarantees for
client-specific relevance estimation or exact subset recovery under partial observations~\cite{mei2024fedmoe,chen2026efficient,zhang2025flexmoe}.
How to reliably allocate experts under strict subset constraints remains largely unexplored.

Restricting clients to only a subset of experts fundamentally changes the MoE learning. 
Under partial expert visibility, routing decisions are no longer updated over the full expert pool, and the observed routing statistics become incomplete and biased. 
Moreover, expert--client relevance evolves throughout training due to evolving model parameters and heterogeneous client objectives. 
Consequently, the server must identify suitable experts for each client using only partial and temporally evolving observations, while avoiding premature specialization or allocation collapse during early training.

To address these challenges, we propose \textbf{$S^3$-MoE} (\textbf{S}erver-\textbf{S}ide \textbf{S}parse \textbf{MoE}) framework in which the server selects and assigns a subset of experts to each client at every communication round, enabling local training under \emph{partial} expert visibility. We formulate expert allocation as a \emph{latent relevance estimation} problem under partial observations and iteratively refine subset
assignment through an exploit--explore strategy. Exploration decays from broad initial coverage to a small persistent level, and relevance is estimated over a sliding window, limiting the influence of unreliable early observations and tracking evolving client--expert relevance. We also present theoretical analysis and experiments for $S^3$-MoE, showing that it recovers task-dependent expert specialization, mitigates expert collapse associated with correlated local Top-$K$ expert selection, and achieves competitive performance under strict expert-subset constraints. Our main contributions are:

\begin{itemize}

\item \textit{Server-side relevance learning under partial expert visibility.}
We study federated MoE with strict subset constraints.
We formulate server-side expert assignment as an online relevance-learning problem under censored and time-varying expert observations. Unlike prior methods based on cross-client usage similarity or directly observed accuracy/loss scores, our formulation learns latent client--expert relevance from partial and time-varying observations.

\item \textit{Relevance estimation and subset recovery algorithm.}
We propose $S^3$-MoE, which combines propensity-corrected sliding-window relevance estimation with an exploit--explore assignment strategy to recover task-dependent top-$K$ expert subsets from partial observations while tracking evolving client--expert preferences. We also establish relevance-estimation error bounds and exact top-$K$ subset-recovery guarantees.

\item \textit{Empirical validation.}
On CIFAR-100, $S^3$-MoE recovers task-dependent client--expert preferences
and improves early-round accuracy and convergence over Random-5.
It achieves higher mean final accuracy than FLEX-MoE-ACC and Full-20 FedAvg,
and remains within 0.60 percentage points of full-visibility $A^3$S-MoE
while synchronizing only five of 20 experts per round.

\end{itemize}

\textbf{Related Work.}
For federated LLM fine-tuning, some methods use LoRA-based adaptation to reduce training overhead, but still require loading the full model locally~\cite{hu2021lora,fedalora2025}. Split learning reduces local memory usage at the cost of increased communication overhead due to repeated transmission of intermediate activations~\cite{liu2024splitfl}.

Recently, MoE has been explored in FL and falls into two relevant groups: full-visibility and partial-visibility of experts. Full-visibility methods retain the complete expert pool locally and apply sparse routing or activation-aware aggregation after local selection. They mainly focus on personalization and heterogeneity rather than local resource constraints~\cite{reisser2021fedmix,dun2023fedjets,yi2024pfedmoe,luo2024pfedmoap}. Resource-aware extensions such as UB-SMoE and $A^3$S-MoE adapt expert activation or aggregation to client capabilities, but still assume full local access to the expert pool~\cite{tran2025ubsm,a3smoe2025}. However, they do not consider strict memory limitations. To address memory constraints, recent works consider loading only a subset of experts. FedMoE constructs personalized sub-MoEs under client memory constraints. It first collects full-pool activation statistics from resource-rich clients to initialize sub-MoEs for constrained clients, and subsequently adds or removes experts using similarity-based recommendations when local performance stops improving~\cite{mei2024fedmoe}. FLEX-MoE operationalizes this design by maximizing accuracy/loss score subject to load-balancing constraints, primarily preventing expert collapse and under-utilization~\cite{zhang2025flexmoe}. However, $S^3$-MoE focuses on identifying the experts most \emph{relevant} to each client's data under a given local expert-access budget. Unlike FedMoE~\cite{mei2024fedmoe}, $S^3$-MoE does not requires resource-rich reference clients with full expert visibility. Unlike FLEX-MoE~\cite{zhang2025flexmoe}, which emphasizes system-wide load balance, $S^3$-MoE uses explicit exploration and propensity-corrected relevance estimation to recover \emph{personalized} expert subsets. In addition, $S^3$-MoE is the first federated MoE framework to provide relevance-estimation error bounds and exact top-$K$ recovery guarantees under partial expert visibility.

\section{System Model and Problem Formulation}

\subsection{Federated MoE System Model}
\label{sec:system_model}

We consider a federated MoE system consisting of a central server and $N$ heterogeneous clients $\mathcal I = \{1,\dots,N\}$. The server maintains a shared pool of $M$ experts, denoted by 
$\mathcal E = \{\mathcal E_m(\cdot; \mathbf W_m^e)\}_{m=1}^M$. Each expert $\mathcal E_m: \mathbb R^{d_x} \to \mathbb R^{d_h}$ is parameterized by $\mathbf W_m^e$. We denote the collection of expert parameters by $\mathbf W^e = \{\mathbf W_m^e\}_{m=1}^M$.

\subsubsection{Subset-based expert allocation}
At communication round $t$, the server assigns to each client $i \in \mathcal I$ a subset of experts $\mathcal S_i^t \subseteq \{1,\dots,M\}$ and transmits the corresponding parameters $\mathbf W_i^{e,t} = \{\mathbf W_m^e\}_{m\in \mathcal S_i^t}$. The subset size $s_i^t$ may vary across clients due to system constraints.

\subsubsection{Client-side MoE computation}
Given the subset $\mathcal S_i^t$, client $i$ performs local training using only the received expert parameters. Let $x \in \mathbb R^{d_x}$ denote an input representation. The client applies a routing function $r_i(\cdot; \mathbf W_i^r): \mathbb R^{d_x} \to \mathbb R^{M},$
which produces routing logits over all experts. In practice, routing is restricted to the assigned subset by applying a masked softmax: $w_{i,m}(x)=0$ for $m \notin \mathcal S_i^t,$ and
\begin{equation}
w_{i,m}(x)
=
\frac{
\exp(r_{i,m}(x; \mathbf W_i^r))
}{
\sum_{m' \in \mathcal S_i^t}
\exp(r_{i,m'}(x; \mathbf W_i^r))
},
\quad m \in \mathcal S_i^t
\label{eq:routing}
\end{equation}
 The representation is $h_i(x)
=
\sum_{m \in \mathcal S_i^t}
w_{i,m}(x)\,\mathcal E_m(x; \mathbf W_m^e)
\label{eq:moe_output}.$

\subsubsection{Local personalization}
Each client maintains a local adaptation module parameterized by $\mathbf W_i^{\mathrm{loc}}$ for task-specific refinement. This module can be instantiated using lightweight techniques such as LoRA or local adapters. The parameters $\mathbf W_i^{\mathrm{loc}}$ remain on the device, enabling personalization while preserving data privacy. The final client model is given by
\[
f_i(x) = f_i^{\mathrm{loc}}(h_i(x); \mathbf W_i^{\mathrm{loc}}).
\]
We emphasize that this component is not the focus of our analysis; our theoretical and algorithmic developments concentrate on the server-side allocation of shared experts.

\subsection{Federated Learning Objective}\label{sec:problem_formulation}

We consider an FL setting in which clients have heterogeneous data distributions and potentially different tasks. Each client $i \in \mathcal{I}$ draws data from a local distribution $\mathcal{D}_i$, and $\ell(\cdot,\cdot)$ denotes the local loss function.
We partition the model parameters into three components:
(i) shared expert parameters $\mathbf{W}^{e} = \{\mathbf{W}^{e}_m\}_{m=1}^M$;
(ii) client-specific routing parameters $\{\mathbf{W}_i^{r}\}_{i=1}^N$;
and (iii) client-private personalization parameters
$\{\mathbf{W}_i^{\mathrm{loc}}\}_{i=1}^N$.
The overall FL objective is formulated as
\[\min_{\mathbf{W}^{e}, \{\mathbf{W}_i^{r}, \mathbf{W}_i^{\mathrm{loc}}\}_{i=1}^N}
\sum_{i=1}^N 
\mathbb{E}_{\mathcal{D}_i}
\Big[
\ell\big(
f_i(x; \mathbf{W}^{e}, \mathbf{W}_i^{r}, \mathbf{W}_i^{\mathrm{loc}}),\, y
\big)
\Big].\]
Subset-based expert allocation introduces several challenges. 
1) Due to client heterogeneity, different clients may require different expert subsets, making expert selection nontrivial. 
2) Early-stage allocation suffers from limited information, which may lead to suboptimal assignments and allocation collapse. 
3) The learning process is non-stationary: expert parameters and routing evolve over time, while relevance must be estimated from partial observations, making consistent selection difficult.

\section{Server-side Sparse Assignment Framework}

We propose $S^3$-MoE, an algorithm to address the challenges introduced in Section~\ref{sec:problem_formulation}. 
The key idea is to maintain a client-specific relevance estimate for each expert and iteratively refine subset assignment through an exploit--explore mechanism.

\subsubsection{Stage 1: Server-side Expert Assignment}

To handle client heterogeneity, the server maintains relevance estimates $\hat P_{i,m}^{\,t}$ for each client--expert pair, which serve as proxies for latent expert relevance under partial observations.

To mitigate unreliable estimates in early stages, we adopt an exploit--explore strategy:
(i) \textbf{Exploitation:} select the top-$k_i^{\mathrm{exp}}$ experts with the highest $\hat P_{i,m}^{\,t}$;
(ii) \textbf{Exploration:} sample the remaining $s_i - k_i^{\mathrm{exp}}$ experts using a usage-balanced random policy to ensure nonzero selection probability. The assigned subset is given by
\begin{equation}
\mathcal S_i^t
=
\mathcal S_{i,\mathrm{exp}}^t
\cup
\mathcal S_{i,\mathrm{expl}}^t,
\mathcal S_{i,\mathrm{exp}}^t
=
\mathrm{TopK}\bigl(\{\hat P_{i,m}^{\,t}\}_{m=1}^M,\, k_i^{\mathrm{exp}}\bigr)
\label{eq:assignment}
\end{equation}
Importantly, the size of the exploration set, $|\mathcal S_{i,\mathrm{expl}}^t|$, changes during training over rounds $t \in {0,\ldots,T-1}$ according to three stages.
\textit{I: Balanced coverage.}
All available slots are used for exploration, i.e., $|\mathcal S_{i,\mathrm{expl}}^t|t=s_i$. Instead of sampling experts uniformly and independently, the
server prioritizes under-observed experts so that every expert is
assigned a prescribed number of times.
\textit{II: Exploration decay.}
After the covering all of experts, $|\mathcal S_{i,\mathrm{expl}}^t|$ gradually decreases
according to a predefined decay schedule. The exploitative slots are filled using the experts with the largest relevance estimates, whereas the remaining slots maintain
usage-balanced exploration.
\textit{III: Stable tracking.}
After the decay stage, the server retains a minimum number
$|\mathcal S_{i,\mathrm{expl}}^t|>0$. This residual exploration preserves
nonzero assignment probabilities and allows the estimator to adapt to
changes in client data distributions or expert relevance.

\subsubsection{Stage 2: Local Update and Server Aggregation}

Client $i$ receives $\mathcal S_i^t$ and performs local updates on 
$\{\mathbf W_m^{e,t}\}_{m\in\mathcal S_i^t}$, $\mathbf W_i^{r,t}$, and $\mathbf W_i^{\mathrm{loc},t}$. 
After local training, client $i$ computes
\begin{equation}
\Delta \mathbf W_{i,m}^{e,t}
=
\mathbf W_{i,m}^{e,t+1} - \mathbf W_m^{e,t},
\quad m\in\mathcal S_i^t
\label{eq:local_update}
\end{equation}
and relevance statistics $\hat r_{i,m}^t$, defined as the \textit{average routing weight} in \eqref{eq:routing} over local samples. The client sends $\{\Delta \mathbf W_{i,m}^{e,t}\}$ and $\{\hat r_{i,m}^t\}$ to the server.
The server aggregates updates as
\begin{equation}
\mathbf W_m^{e,t+1}
=
\mathbf W_m^{e,t}
+
\mathrm{Agg}\bigl(\{\Delta \mathbf W_{i,m}^{e,t}\}_{i\in\mathcal K_m^t}\bigr),
\label{eq:aggregation}
\end{equation}
where $\mathcal K_m^t = \{i: m\in\mathcal S_i^t\}$. The relevance estimates are updated using the observations from the most
recent $\min\{W_t,t+1\}$ rounds, with the window defined as
$W_t=\{\max(0,t-W_t+1),\ldots,t\}$ and the conditional
assignment probability denoted by $\rho_{i,m}^{\,\tau}$:
\begin{equation}
\hat P_{i,m}^{\,t+1}
=
\frac{1}{| W_t|}
\sum_{\tau\in W_t}
\frac{
\mathbf 1\{m\in\mathcal S_i^\tau\}
\hat r_{i,m}^{\,\tau}
}{
\rho_{i,m}^{\,\tau}
},
\label{eq:relevance_update}
\end{equation}

\begin{algorithm}[ht]
\caption{Server-Side Sparse Expert Assignment}
\label{alg:main}
\begin{algorithmic}[1]
\Require Experts $\{\mathbf W_m^{e,0}\}_{m=1}^M$, clients $\mathcal I$, subset sizes $\{s_i\}$, window $W_t$, Initial relevance $\hat P_{i,m}^{\,0}$

\For{$t=0,1,\dots,T-1$}

\State \textbf{Server: Assignment}
\For{each $i\in\mathcal I$}
    \State $\mathcal S_i^t \gets \mathcal S_{i,\mathrm{exp}}^t \cup \mathcal S_{i,\mathrm{expl}}^t$ using \eqref{eq:assignment}
    \State Send $\mathcal S_i^t$ and $\{\mathbf W_m^{e,t}\}_{m\in\mathcal S_i^t}$ to client $i$
\EndFor

\State \textbf{Client: Local update}
\For{each $i\in\mathcal I$}
    \State Update $\{\mathbf W_m^{e,t}\}_{m\in\mathcal S_i^t},\mathbf W_i^{r,t},\mathbf W_i^{\mathrm{loc},t}$
    \State Compute $\{\Delta \mathbf W_{i,m}^{e,t}\}$ using \eqref{eq:local_update} and $\{\hat r_{i,m}^t\}_{m\in\mathcal S_i^t}$
    \State Send  $\{\Delta \mathbf W_{i,m}^{e,t}, \hat r_{i,m}^t\}_{m\in\mathcal S_i^t}$ to the server
\EndFor
\State \textbf{Server: Aggregation and update}
\For{$m=1,\dots,M$}
\State Update $\mathbf W_m^{e,t+1}$ via aggregation in \eqref{eq:aggregation}
\State Compute $\hat P_{i,m}^{\,t+1}$ using \eqref{eq:relevance_update}
\EndFor
\EndFor
\end{algorithmic}
\end{algorithm}

\section{Theoretical Results}
\label{sec:theory}

In this section, we analyze the estimation error of the proposed relevance estimator $\hat P_{i,m}^{\,t}$ in Algorithm~\ref{alg:main}, and establish conditions under which the optimal top-$K$ expert subset can be recovered. Importantly, under subset-restricted routing, the observed routing weights are normalized over a partial expert set and are therefore not directly comparable to full-visibility routing. 
Instead of recovering full-pool routing weights, we focus on estimating a latent \emph{relevance} quantity that captures the expected usefulness of each expert under partial observations. 
Our goal is to accurately estimate these \textit{relevance scores} and recover the corresponding \textit{top-$K$ subset}.
We define the latent relevance of expert $m$ for
client $i$ as $P_{i,m}^{*,t}
:=
P_{i,m}^*(\mathbf W_i^{r,t})
=
\mathbb E\!\left[
\hat r_{i,m}^{\,t}
\mid
m\in\mathcal S_i^t,\mathcal F_{t-1}
\right].$
Throughout the paper, $\|\cdot\|$ denotes the $\ell_2$ norm.

\begin{assumption}[Lipschitz continuity]\label{as:L_continuity}
For all $i,m$ and all $\mathbf W_i^{r},\mathbf W_i^{r'}$, $|P_{i,m}^*(\mathbf W_i^r)-P_{i,m}^*(\mathbf W_i^{r'})|
\le
L_P \|\mathbf  W_i^{r}-\mathbf  W_i^{r'}\|.$
\end{assumption}

\begin{assumption}[Minimum assignment probability]\label{as:min_pro}
There exists $\rho_{\min}>0$ such that
$\rho_{i,m}^{\,t} := \Pr(m\in \mathcal S_i^t \mid \mathcal F_{t-1}) \ge \rho_{\min},
\forall i,m,t.$
\end{assumption}

Assumption~\ref{as:L_continuity} is mild because the router is typically
a lightweight neural network with bounded parameters and Lipschitz
activations; hence its routing weights, and their conditional
expectations, are locally Lipschitz in $\mathbf W_i^r$. Assumption 2 is naturally satisfied under any exploration policy that assigns each expert with nonzero probability.

\begin{theorem}[Relevance estimation error bound]
\label{thm:main_error}
Suppose Assumptions~\ref{as:L_continuity},\ref{as:min_pro}  hold. Then for any $\delta\in(0,1)$, with probability at least $1-\delta$,
{\small
\[    \bigl|
\hat P_{i,m}^{\,t} - P_{i,m}^{*,t}
\bigr|\le
\underbrace{
\sqrt{\frac{2\log(2/\delta)}{W_t \rho_{\min}^2}}
}_{\text{sampling err}}+
\underbrace{
L_P \sum_{j=t-W_t+1}^{t-1}
\left\|
\mathbf W_i^{r,j+1}
-
\mathbf W_i^{r,j}
\right\|
}_{\text{drift err}}.
\]
}
\end{theorem}
To simplify notation, we define
\[
\varepsilon_t
:=
\frac{1}{\rho_{\min}}
\sqrt{\frac{2\log(2M/\delta)}{W_t}}
+
L_P \sum_{j=t-W_t+1}^{t-1}
\left\|
\mathbf W_i^{r,j+1}
-
\mathbf W_i^{r,j}
\right\|.
\]
Here the $\log(2M/\delta)$ term follows from a union bound over all $M$ experts.

\begin{theorem}[Exact top-$K$ recovery]
\label{thm:topk_recovery}
Let $\pi_t$ be a permutation of $\{1,\dots,M\}$ such that 
$P_{i,\pi_t(1)}^{*,t} \ge \cdots \ge P_{i,\pi_t(M)}^{*,t}$, and define 
$S_i^{*,t} := \{\pi_t(1),\dots,\pi_t(K)\}$. 
Let $\hat S_i^{t}$ be the top-$K$ set induced by $\hat P_{i,m}^{\,t}$. 
If $P_{i,\pi_t(K)}^{*,t} - P_{i,\pi_t(K+1)}^{*,t} > 2\varepsilon_t,$
then with probability at least $1-\delta$, $\hat S_i^t = S_i^{*,t}$.
\end{theorem}

\textbf{Proof of Theorem~\ref{thm:main_error}.}
$\bar P_{i,m}^{*,(t,W_t)} := \frac{1}{W_t}\sum_{\tau=t-W_t+1}^{t} P_{i,m}^{*,\tau}$. Then
$
|\hat P_{i,m}^{\,t}-P_{i,m}^{*,t}|
\le
|\hat P_{i,m}^{\,t}-\bar P_{i,m}^{*,(t,W_t)}|
+
|\bar P_{i,m}^{*,(t,W_t)}-P_{i,m}^{*,t}|.
$ We call the first term the sampling error and the second term the drift error. \textit{Step 1 (Sampling error).}
Since
$P_{i,m}^{*,\tau}=\mathbb E[\hat r_{i,m}^{\,\tau}\mid m\in\mathcal S_i^\tau,\mathcal F_{\tau-1}]$,
the estimator is conditionally unbiased, yielding a martingale difference sequence. By Azuma--Hoeffding~\cite{azuma1967weighted}, with probability at least $1-\delta$,
$
|\hat P_{i,m}^{\,t}-\bar P_{i,m}^{*,(t,W_t)}|
\le
\frac{1}{\rho_{\min}}\sqrt{\frac{2\log(2/\delta)}{W_t}}.
$ 
\textit{Step 2 (Drift error).}
The second term captures temporal variation of relevance:
$
|\bar P_{i,m}^{*,(t,W_t)}-P_{i,m}^{*,t}|
\le
\frac{1}{W_t}\sum_{\tau=t-W_t+1}^{t}
|P_{i,m}^{*,\tau}-P_{i,m}^{*,t}|.
$
By Assumption~\ref{as:L_continuity},
$
|P_{i,m}^{*,\tau}-P_{i,m}^{*,t}|
\le
L_P\|\mathbf W_i^{r,\tau}-\mathbf W_i^{r,t}\|,
$
and by telescoping,
$
\|\mathbf W_i^{r,\tau}-\mathbf W_i^{r,t}\|
\le
\sum_{j=\tau}^{t-1}\|\mathbf W_i^{r,j+1}-\mathbf W_i^{r,j}\|.
$
Combining these gives
$
|\bar P_{i,m}^{*,(t,W_t)}-P_{i,m}^{*,t}|
\le
L_P\sum_{j=t-W_t+1}^{t-1}\|\mathbf W_i^{r,j+1}-\mathbf W_i^{r,j}\|.
$

\textbf{Proof of Theorem~\ref{thm:topk_recovery}.}
By Theorem~\ref{thm:main_error} and a union bound over all experts, $\|\hat P_i^{\,t}-P_i^{*,t}\|_\infty \le \varepsilon_t$ with probability at least $1-\delta$. For any $m \in S_i^{*,t}$ and $m' \notin S_i^{*,t}$, we have $P_{i,m}^{*,t} \ge P_{i,(K)}^{*,t},
P_{i,m'}^{*,t} \le P_{i,(K+1)}^{*,t}.$
Using the uniform bound, $\hat P_{i,m}^{\,t} \ge P_{i,m}^{*,t} - \varepsilon_t,
\hat P_{i,m'}^{\,t} \le P_{i,m'}^{*,t} + \varepsilon_t,$
which implies
$\hat P_{i,m}^{\,t} - \hat P_{i,m'}^{\,t}
\ge
P_{i,(K)}^{*,t} - P_{i,(K+1)}^{*,t} - 2\varepsilon_t.$
If the gap satisfies
$P_{i,(K)}^{*,t} - P_{i,(K+1)}^{*,t} > 2\varepsilon_t,$
then $\hat P_{i,m}^{\,t} > \hat P_{i,m'}^{\,t}$ for all such pairs, implying $\hat S_i^t = S_i^{*,t}$.

\textbf{Discussion.}
In Theorem~\ref{thm:main_error} the estimation error consists of two components: a \emph{sampling error} and a \emph{drift error}. 
The sampling error is due to partial observation and is controlled by the assignment probability $\rho_{\min}$ and the window size $W_t$, where larger $\rho_{\min}$ improves coverage and larger $W_t$ reduces variance. 
The drift error reflects changes in relevance caused by updates
to the local router parameters $\mathbf W_i^{r,t}$. As the differences
between successive router parameters decrease, the drift error decreases. The router learns its routing policy but remains coupled
with the experts through joint optimization: expert updates affect
the local training loss and, consequently, subsequent router gradients.
Thus, stabilization of both expert and router updates leads to more stable relevance estimation.

Theorem~\ref{thm:topk_recovery} shows that correct top-$K$ recovery holds when the relevance gap exceeds the estimation error. The analysis further implies partial recovery: the top-$K'$ experts can still be correctly identified when the gap condition $P_{i,\pi_t(K')}^{*,t} - P_{i,\pi_t(K+1)}^{*,t} > 2\varepsilon_t$ holds.

\section{Experimental Results}

\begin{table*}[t]
\centering
\scriptsize
\setlength{\tabcolsep}{2pt}
\renewcommand{\arraystretch}{0.92}
\caption{Accuracy and communication of methods on CIFAR-100
(mean $\pm$ standard deviation over 10 seeds).}
\begin{tabular*}{\textwidth}{@{\extracolsep{\fill}}clclccc@{}}
\hline
Experts & Method & Router & Synchronized scope & Acc. (\%) &
Client/round (MiB) & Total (GiB) \\
\hline
\multirow{5}{*}{5}
& \textbf{$S^3$-MoE} & Local & 5 experts
& $\mathbf{53.05\pm0.39}$ & \textbf{5.99} & \textbf{5.85} \\
& $S^3$-MoE-Greedy & Local & 5 experts
& $53.02\pm0.42$ & \textbf{5.99} & \textbf{5.85} \\
& Random-5 & Local & 5 experts
& $47.53\pm0.57$ & \textbf{5.99} & \textbf{5.85} \\
& FLEX-MoE-ACC (personalized) & Local & 5 experts
& $52.82\pm0.48$ & \textbf{5.99} & \textbf{5.85} \\
& FLEX-MoE-ACC & Local & Backbone + 5 experts
& $46.93\pm0.23$ & 72.83 & 71.12 \\
\hline
\multirow{5}{*}{20}
& $A^3$S-MoE (personalized) & Local & 20 experts
& $\mathbf{53.65\pm0.49}$ & 23.96 & 23.40 \\
& $A^3$S-MoE (global-router) & Global & 20 experts + router
& $53.06\pm0.33$ & 24.04 & 23.48 \\
& $A^3$S-MoE (original) & Global & All parameters
& $35.02\pm0.59$ & 90.88 & 88.75 \\
& Full-20 FedAvg (personalized) & Local & 20 experts
& $51.74\pm0.39$ & 23.96 & 23.40 \\
& Full-20 FedAvg (original) & Global & All parameters
& $37.46\pm0.61$ & 90.88 & 88.75 \\
\hline
\end{tabular*}
\label{tab:accuracy_communication}
\end{table*}

\textit{Experimental setup.}
We use CIFAR-100~\cite{krizhevsky2009learning}, which contains 100 fine
classes grouped into 20 coarse superclasses. 10 clients form 5 strongly
non-IID task groups, with 2 clients and 4 coarse classes per group. Each
client contains 5,000 training samples and 1,000 test samples. Each client performs 12 mini-batch
updates with batch size 128 in every communication round, corresponding to
1,536 processed training examples per client per round. The model
consists of a CNN feature extractor and projection parameterized by
$\mathbf{W}_i^{\mathrm{loc}}$, a router parameterized by
$\mathbf{W}_i^r$, and one MoE layer containing $M=20$ fully connected
experts with hidden dimension 256. Capacity-constrained methods receive five
experts and route densely over the assigned subset, whereas Full-20 FedAvg and
$A^3$SMoE receive all 20 experts and activate a local Top-5 subset. Experts
are initialized from a task-specialized checkpoint whose client--expert
correspondence is hidden from the server. All methods are trained for 100 global communication rounds using AdamW with
learning rate $3\times10^{-4}$ and weight decay $10^{-4}$. Heatmaps use seed 0, whereas
Table~\ref{tab:accuracy_communication} reports the mean and standard deviation
over seeds 0--9.

\textit{Baselines and controlled variants.}
Random-5, $S^3$-MoE, and $S^3$-MoE-Greedy each allocate five experts and
differ only in their server-side assignment strategies. Random-5 randomly
samples five experts. $S^3$-MoE and $S^3$-MoE-Greedy use the same relevance
estimation strategy, but $S^3$-MoE retains one exploration expert, whereas
$S^3$-MoE-Greedy uses no exploration experts after the eighth iteration.
FLEX-MoE-ACC~\cite{zhang2025flexmoe} performs resource-balanced assignment
using accuracy feedback, allocates five experts, and aggregates the shared
backbone during training. Full-20 FedAvg~\cite{mcmahan2017communication} and
$A^3$SMoE~\cite{a3smoe2025} synchronize all 20 experts. Unlike FedAvg,
$A^3$SMoE uses activation-aware expert aggregation to reduce expert collapse.
We evaluate personalized parameter-sharing variants to separate the effects of
expert assignment and personalization; only the selected expert subset or full
expert bank is synchronized.

\begin{figure}[t]
    \centering
    \includegraphics[width=0.95\linewidth]{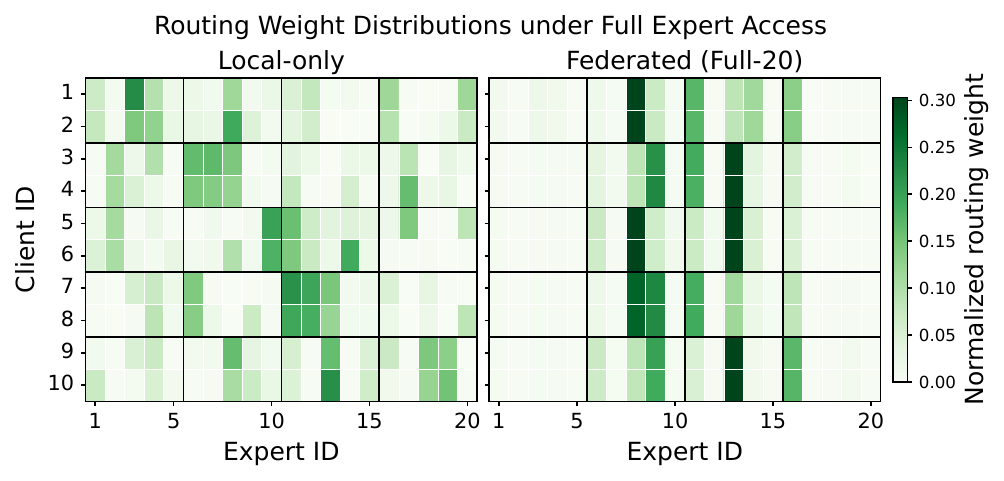}
    \caption{
  Routing weights under full access.
Local training shows specialization, while federated training preserves alignment but reduces diversity.
    }
    \label{fig:full_access}
\end{figure}

\begin{figure}[t]
    \centering
    \includegraphics[width=0.95\linewidth]{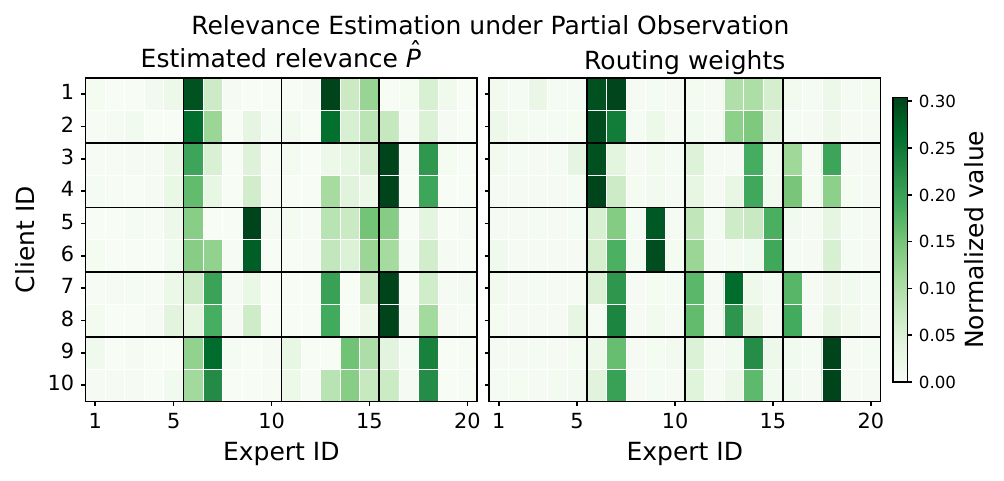}
    \caption{
    Estimated relevance $\hat{P}$ and corresponding routing weights under partial observation.
    }
    \label{fig:phat_estimation}
\end{figure}

\textbf{Experimental Results and Discussion.}
\textit{Server-side vs.\ local expert selection.} Figure~\ref{fig:full_access} shows routing weight distributions under local-only and federated expert fine-tuning. Under local-only training, expert usage exhibits clear task-dependent specialization, where clients with similar data select similar experts, and this selection matches the pretrained checkpoint. Under FL, such preferences are partially preserved, but expert usage becomes less diverse due to aggregation-induced collapse toward dominant experts.
Figure~\ref{fig:phat_estimation} shows that, under partial expert
visibility, the estimated relevance $\hat P$ recovers group-dependent
client--expert preferences and produces more distributed expert
exposure than the Full-20 FedAvg setting in Figure~\ref{fig:full_access}.
With Full-20 FedAvg, clients \textit{independently} perform local Top-5 selection from
the complete expert pool. Consequently, their local routers may
repeatedly select the same experts that become dominant during early
training, concentrating updates on a small subset and causing expert
collapse. In contrast, $S^3$-MoE coordinates expert exposure at the
server before local training, preventing dominant experts from being
repeatedly assigned across clients. 

\textit{Accuracy and communication comparison.}
Table~\ref{tab:accuracy_communication} compares methods with five- and
20-expert access. Under the five-expert budget, $S^3$-MoE achieves the highest
mean accuracy. $S^3$-MoE-Greedy performs similarly but has slightly lower
accuracy and higher variance. $S^3$-MoE also outperforms Random-5 and both
FLEX-MoE-ACC variants. Across FLEX-MoE, $A^3$S-MoE, and Full-20 FedAvg,
variants retaining a local backbone and router generally perform better,
suggesting that personalization is beneficial under strongly non-IID data.
Moreover, $S^3$-MoE is only 0.60 percentage points below the full-visibility
$A^3$S-MoE baseline while reducing communication by 75\%. Compared with
personalized Full-20 FedAvg, it improves accuracy by 1.31 percentage points
while achieving the same communication reduction.

\section{Conclusion and Future Work}
We studied federated MoE settings in which clients access only a limited subset
of experts, formulated server-side subset allocation under partial expert
visibility, and proposed $S^3$-MoE to recover task-dependent top-$K$ expert
subsets. Theoretical analysis provides relevance-estimation and exact top-$K$
recovery guarantees, while experiments demonstrate effective specialization
and accuracy comparable to full-visibility methods at substantially lower
communication cost. Future work will extend $S^3$-MoE to multi-layer MoE LLMs
and investigate adaptive exploration--exploitation strategies for evolving
expert relevance.

\section*{Acknowledgment}
This work was supported in part by the National Science Foundation under Grant CNS-2112471.

\bibliographystyle{IEEEtran} %
\bibliography{refs.bib} %

\end{document}